\documentclass{article}
\usepackage{spconf,amsmath,graphicx}

\usepackage{algorithm}
\usepackage{algorithmic}

\usepackage{multirow}
\usepackage{amsthm}
\usepackage{booktabs}
\usepackage{epsfig}
\usepackage{subfigure}
\usepackage{subfig}
\usepackage{amssymb}
\usepackage{pifont}

\usepackage{newfloat}
\usepackage{listings}

\title{Detecting Out-of-distribution Examples via Class-conditional Impressions Reappearing}
\name{Jinggang Chen$^{\dag\ddag}$, Xiaoyang Qu$^{\ddag*}$, Junjie Li$^{\dag}$, Jianzong Wang$^{\ddag}$, Jiguang Wan$^{\dag}$, Jing Xiao$^{\ddag}$\thanks{* Corresponding author: Xiaoyang Qu (e-mail: quxiaoy@gmail.com).}}
\address{$^{\dag}$ Huazhong University of Science and Technology, China \quad$^{\ddag}$ Ping An Technology (Shenzhen) Co., Ltd.}
\begin{document}
%
\maketitle
\begin{abstract}
Out-of-distribution (OOD) detection aims at enhancing standard deep neural networks to distinguish anomalous inputs from original training data. Previous progress has introduced various approaches where the in-distribution training data and even several OOD examples are prerequisites. However, due to privacy and security, auxiliary data tends to be impractical in a real-world scenario. In this paper, we propose a data-free method without training on natural data, called Class-Conditional Impressions Reappearing (C2IR),  which utilizes image impressions from the fixed model to recover class-conditional feature statistics. Based on that, we introduce Integral Probability Metrics to estimate layer-wise class-conditional deviations and obtain layer weights by Measuring Gradient-based Importance (MGI). The experiments verify the effectiveness of our method and indicate that C2IR outperforms other post-hoc methods and reaches comparable performance to the full access (ID and OOD) detection method, especially in the far-OOD dataset (SVHN).   
\end{abstract}
\section{Introduction}
\label{sec:intro}
Out-of-distribution (OOD) detection is crucial to ensuring the reliability and safety of AI applications. While OOD examples from the open-world are prone to induce over-confident predictions, which makes the separation between In-distribution (ID) and OOD data a challenging task.

Previous works have proposed a bunch of approaches to estimate the distribution discrepancy between training data and target OOD examples \cite{MaxConfidenceScore,ma_distance, ODIN,liu2020energy}. Despite the progress, conventional OOD detection methods universally rely on ID examples and even OOD examples for detector training or hyper-parameter tuning. However, as privacy and security matter more nowadays, some original datasets, especially the self-made datasets in the business application field, have become unavailable, and the distribution of OOD examples is unpredictable. Thus, utilizing auxiliary data is impractical in the real-world scenario.

\begin{figure}[!t]
    \centering
	\includegraphics[width=8cm]{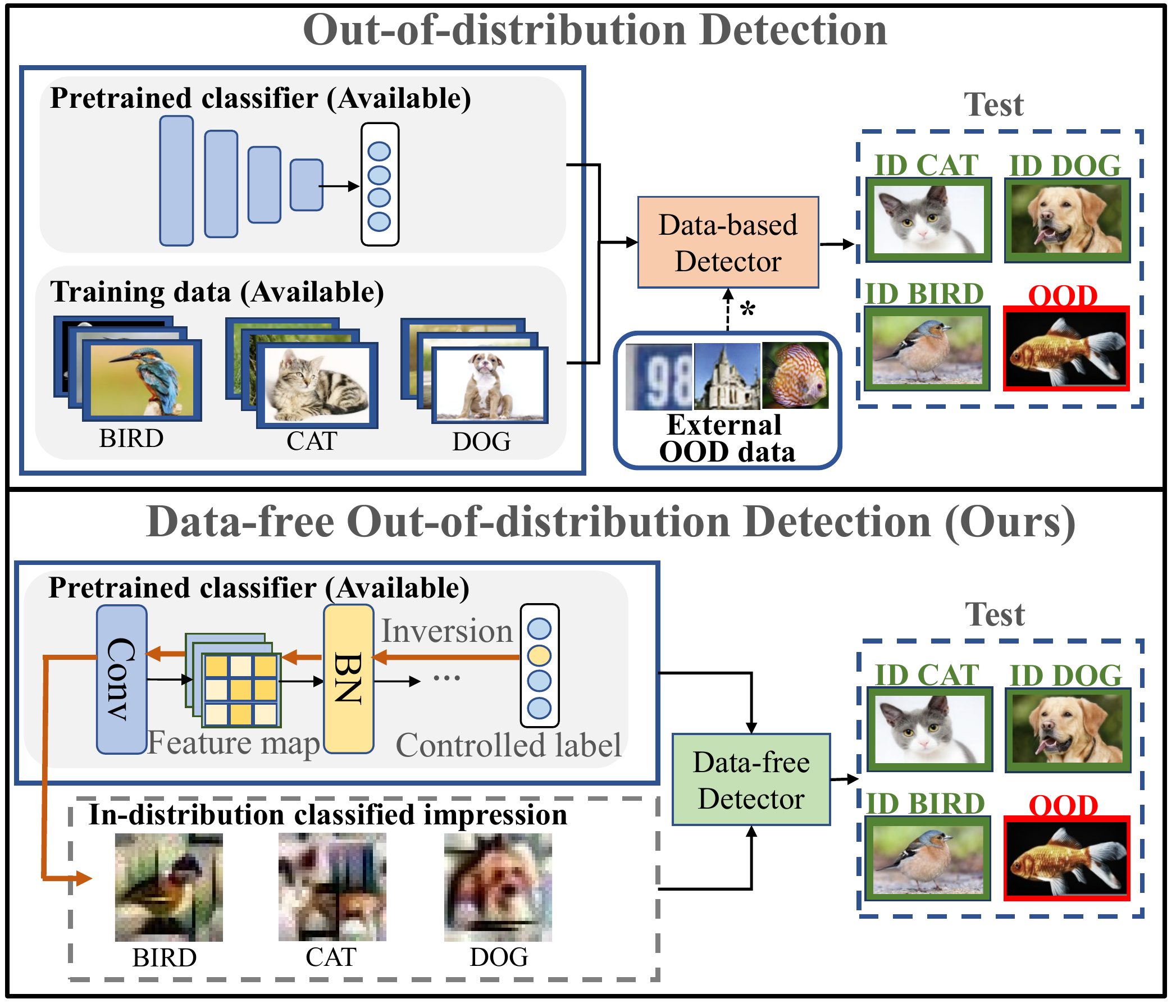}
	\caption{Comparison between typical out-of-distribution detection methods and ours. In the typical methods, both the pretrained model and training data are required. * means several approaches \cite{ma_distance, OE} additionally assume access to OOD data. While our method only requires the fixed classifier and utilizes the synthesized data to detect OOD examples.}
	\label{fig:framework}
   	\vspace{-0.5em}
\end{figure}

In this paper, we propose C2IR, a data-free detection method via Class-Conditional Impressions Reappearing. Several recent studies have shown that pretrained models contain sufficient information about the distribution of training data \cite{React,nmd}. We further exploit more detailed virtual statistics from the pretrained model with the inversion trick \cite{deepinv}. 

Our contributions are as follows:
\begin{itemize}
    \item We propose that the image impressions from the model's BatchNorm layers share similar intermediate feature statistics with in-distribution data. 
    \item Based on the virtual data, we design the layer-wise deviation metrics by utilizing virtual class-conditional activations mean. 
    \item To obtain the layer weights without the support of auxiliary data, we propose an attribution method by Measuring Gradient-based Importance (MGI).
\end{itemize}



\section{PRELIMINARIES AND MOTIVATIONS}
\label{sec:background}
\subsection{Out-of-distribution Detection}
The goal of out-of-distribution (OOD) detection is to distinguish anomalous inputs ($\mathcal{D}_{\text{out}}$) from original training data ($\mathcal{D}_{\text{in}}$) based on a well-trained classifier $f_{\theta}$. This problem can be considered as a binary classification task with a score function $\mathcal{S}(\cdot)$. Formally, given an input sample $x$, the level-set estimation is:
\begin{equation}
    g(x) = \begin{cases}
    \text{out}, &\text{if}\quad\mathcal{S}(x) > \gamma\\
    \text{in}, &\text{if}\quad\mathcal{S}(x) \leq \gamma\\
    \end{cases}  
\end{equation}
In our work, lower scores indicate that the sample $x$ is more likely to be classified as in-distribution (ID) and $\gamma$ represents a threshold for separating the ID and OOD.

\subsection{BatchNorm Statistics for Approximate Estimation}
Under a Gaussian assumption about feature statistics in the network, feature representations at one layer are denoted as $\textbf{Z}\in \mathbb{R}^{d\times d}$. For training distribution $\mathcal{X}$, each $z\in \textbf{Z}$ follows the gaussian distribution $\mathcal{N}(\mu_{\text{in}}, \sigma^{2}_{\text{in}})$.
Normally the statistics of $\mathcal{X}$ are calculated from training samples. For the purposes of estimation efficiency \cite{BN_efficient, nmd} and data privacy \cite{deepinv}, several works propose that running average from the batch normalization (BN) layer can be an alternative to original in-distribution statistics. During training time, BN expectation $\mathbb{E}_{\text{bn}}(z)$ and variance $\text{Var}_{\text{bn}}(z)$ are updated after a batch of samples $\mathcal{T}$:
\begin{equation}
\begin{matrix}
    \mathbb{E}_{\text{bn}}(z)\leftarrow \lambda\mathbb{E}_{\text{bn}}(z) + (1-\lambda)\frac{1}{|\mathcal{T}|}\sum_{x_b\in \mathcal{T}} \mu(x_b) \\
    
    \text{Var}_{\text{bn}}(z)\leftarrow \lambda\text{Var}_{\text{bn}}(z) + (1-\lambda)\frac{1}{|\mathcal{T}|}\sum_{x_b\in \mathcal{T}} \sigma^2(x_b) 
    \end{matrix}
\end{equation}
Therefore the in-distribution statistics are approximately estimated:  $\mathbb{E}_{\text{in}}(z)\approx \mathbb{E}_{\text{bn}}(z), \text{Var}_{\text{in}}(z)\approx \text{Var}_{\text{bn}}(z)$.


\subsection{Motivation}

Feature statistics have been applied in some advanced post-hoc methods and reach striking performances \cite{ma_distance, nmd, Gram}. Their paradigm involves utilizing statistical data from the feature map as a measure of difference and calculating the final score by weighting and summing the scores from each layer. Prior studies obtain the weights via training in in-distribution (ID) images \cite{nmd, Gram} or even additionally estimating on some OOD datasets \cite{ma_distance, mood}. Since training (ID) samples are unreachable in our setting, BN statistics are readily available resources from pretrained model $f_{\theta}$ for data-free OOD detection.

Previous works in the data-free knowledge distillation\cite{deepinv, dafl, qu2021enhancing} prove that teacher-supervised synthesized images are as effective as the natural data in knowledge transferring. A reasonable assumption is that these pseudo samples constrained by the model's BN statistics recall the impression of training data. Therefore, unlike the direct use of BatchNorm mean \cite{nmd}, we consider an inverting and recollecting method can be more appropriate in the data-free scenario.

\section{PROPOSED METHOD}
\label{sec:method}
\subsection{Detection Framework}
The framework of our method detects an anomalous input $x$ by computing layer-wise deviations $\Delta_l(\cdot)$. With the scaling values $\alpha^{c}_{l}$ for class $c$, score function $S(\cdot)$ is formulated as,
\begin{equation}
S(x)=\sum\limits_{l=1}^{L} \alpha^{c}_{l} \Delta_l(x)  
\end{equation}
where class $c$ is determined by maximum softmax probability (MSP) \cite{MaxConfidenceScore}: $c = \mathop{\text{argmax}}_{i\in [1,C]} f_{\theta}(x)_{i}$. Suppose unknown inputs $\{x\}$ from distribution $\mathcal{Q}$ and original training data $\{x_{\text{in}}\}$ from $\mathcal{X}$. The deviation between $\mathcal{X}$ and $\mathcal{Q}$ can be defined following Integral Probability Metrics\cite{IPM}: $\text{sup}_{\psi}(\mathbb{E}_{x\in \mathcal{Q}}[\psi(x)] - \mathbb{E}_{x_{\text{in}}\in \mathcal{X}}[\psi(x_{\text{in}})])$,
where $\psi(\cdot)$ denotes a witness function. Consider the real condition that only a single input $x$ from $\mathcal{Q}$ is measured. Specially, we further estimate the class-conditional deviation by,
\begin{equation}
 \text{sup}_{\psi}(\psi(x) - \mathbb{E}_{x_{\text{in}}\in \mathcal{X}}[\psi(x_{\text{in}})\mid y=c])
\label{IPM}
\end{equation}
As the in-distribution statistics are invisible to our data-free detection method, we use the expectation of virtual images $\{\hat{x}\}$ to be the replacement. The details of layer-wise deviations and scaling values are discussed in the following sections.

\subsection{Reappearing Data Statistics from BN Layers}
Model inversion is initially designed for artistic effects on clean images \cite{DeepDream} and 
has been developed for synthesizing virtual training data in data-free knowledge distillation  \cite{deepinv}.
In this paper, we utilize this technique to excavate prior class-conditional statistics from the model's BatchNorm layers.


\textbf{Data Recovering.}
Given a random noise $\hat{x}$ as input and a label $c$ as target, recovering process aims at optimizing $\hat{x}$ into a pseudo image with discernible visual information by,
\begin{equation}
\begin{aligned}
    \min\limits_{\hat{x}} \underbrace{\mathcal{L}_{\text{CE}}(f_{\theta}(\hat{x}, c))}_{\text{(i)}} &+ \sum\nolimits_l  \| \mu_l(\hat{x}) - \mathbb{E}^{l}_{bn}(z)\|_{2}\\ &+ \underbrace{\sum\nolimits_l  \| \sigma^2_l(\hat{x}) - \mathbb{\text{Var}}^{l}_{bn}(z)\|_{2}}_{\text{(ii)}}
    \label{eq2}
\end{aligned}
\end{equation}
where $\text{(i)}$ represents class prior loss (cross entropy loss) and $\text{(ii)}$ represents feature statistic (mean and variance) divergences between $\hat{x}$ and BN running average at each $l$ layer.

\textbf{Class-conditional Activation Mean.}
The activation units' discrepancy is introduced to differentiate examples from ID and OOD.
For each class $c$, we control the class prior loss and generate the one-class dataset $\mathcal{D}^{c}_{\text{syn}}$. Denotes the $l$-layer activation map for the input $x$ as $A^{l}(x)\in \mathbb{R}^{h\times d_{l}\times d_{l}}$,
where $h$ is the channel size and $d_l$ is the dimension. Given input $x$ and its MSP label $c$, our layer-wise deviations $\Delta_{l}(\cdot)$ transforms the equation\ref{IPM} with the channel-wise linear weighted average C-Avg($
\cdot$):
\begin{equation}
    \begin{aligned}
        \Delta_{l}(x) &= \text{sup}_{\psi}\mid\psi(x) - \frac{1}{|\mathcal{D}^{c}_{\text{syn}}|} \sum\limits_{\hat{x}\in \mathcal{D}^{c}_{\text{syn}}} \psi(\hat{x}) \mid \\
        &= \mid \text{C-Avg}(A^l(x)) - \underbrace{\frac{1}{|\mathcal{D}^{c}_{\text{syn}}|} \sum\limits_{\hat{x}\in \mathcal{D}^{c}_{\text{syn}}} \text{C-Avg}(A^{l} (\hat{x}))}_{\text{(i)}}\mid \\
        &=  \mid\sum\limits_{k=1}^{h}\beta^{c}_{l,k}\cdot\frac{1}{d_{l}^2}\sum\limits_{i=1}^{d_{l}}\sum\limits_{j=1}^{d_{l}} A^{l}(x)_{k,i,j} \\ 
        &\quad \quad- \frac{1}{|\mathcal{D}^{c}_{\text{syn}}|}\sum\limits_{\hat{x}\in \mathcal{D}^{c}_{\text{syn}}}\sum\limits_{k=1}^{h}\beta^{c}_{l,k}\cdot\frac{1}{d_{l}^2}\sum\limits_{i=1}^{d_{l}}\sum\limits_{j=1}^{d_{l}} A^{l}(\hat{x})_{k,i,j}\mid
    \end{aligned}
\end{equation}
where $\beta^{c}_{l,k}$ denotes the k-th channel weight for class $c$ and $\text{(i)}$ represents the empirical $\overline{\text{C-Avg}}_{c}$ from $\mathcal{D}_{\text{syn}}^{c}$.
\vspace{-1.0em}
\subsection{Measuring Gradient-based Importance (MGI)}
The scaling values $\alpha^{c}_{l}, \beta^{c}_{l,k}$ greatly influence the performance of OOD detection. In our data-free method, the process of image evolution naturally provides an observation for the weights of neuron importance. 

\textbf{Channel-wise Gradient Average.}
Let $y^c$ be the model output score for class $c$. We introduce the gradient-based attribution method \cite{channel_mean} to compute the gradient average $\overline{w}^{c}_{l,k}$ as the weight of $k$-th channel activation map at $l$-th layer.
\begin{equation}
\centering
\begin{aligned}
    \overline{w}^{c}_{l,k}=\frac{1}{|\mathcal{D}^c_{\text{syn}}|}\sum\limits_{\hat{x}\in \mathcal{D}^{c}_{\text{syn}}}\frac{1}{d_{l}^{2}}\sum\limits_{i=1}^{d_{l}}\sum\limits_{j=1}^{d_{l}}\frac{\partial y^{c}}{\partial A^{l}(\hat{x})_{k,i,j}}
\end{aligned}
\end{equation}

\textbf{Layer-wise Activation Mean Sensitivity.} Denotes the image sequence during the optimizing procedure with $T$ iterations as $\{\hat{x}_0, \hat{x}_1, ... \hat{x}_T\}$. Compared with $\hat{x}_T$ containing useful in-distribution features that induce high-confidence prediction \cite{features_not_bugs} for target class $c$, random noise $\hat{x}_0$ is considered to be a baseline without any prior information. Based on the variations of their feature statistics, we define the $l$-layer activations mean sensitivity for $t$-iteration image $\hat{x}_t$ as,
\begin{equation}
\begin{aligned}
    \delta_{l}(\hat{x}_{t}) &= \frac{1}{\Delta y_t^c}\cdot \frac{\partial y_t^c}{\partial \text{C-Avg}(A^l(\hat{x}_{t}))} \\
    &= \frac{1}{y^c_t-y^c_{t-1}}\cdot \sum\limits_{k=1}^{h} \beta^{c}_{l,k}\frac{1}{d_l^{2}}\sum\limits_{i=1}^{d_l}\sum\limits_{j=1}^{d_l} \frac{\partial y^c_{t}}{\partial A^{l}(\hat{x}_{t})_{k,i,j}}
\end{aligned}
\end{equation}
This is, a larger variation of feature statistics during the evolution of $c$-class output score $y^c_t$ indicates that this layer is more sensitive to the label shifts on inputs. In other words, the layer's magnitudes can better differentiate abnormal inputs. Therefore, we use the relative sensitivity $\delta_{l}(\hat{x}_t)$ to measure the layer-contribution for OOD detection. The total $\delta_{l}(\hat{x})$ and empirical $\overline{\delta}^c_{l}$ for $\mathcal{D}^c_{\text{syn}}$ are computed by,
\begin{equation}
    \begin{aligned}
        \delta_{l}(\hat{x}) = \frac{1}{T}\sum\limits_{t=1}^{T} \delta_{l}(\hat{x}_{t}), \overline{\delta}^c_{l} = \frac{1}{|\mathcal{D}^{c}_{syn}|}\sum\limits_{\hat{x}\in \mathcal{D}^c_{\text{syn}}} \delta_{l}(\hat{x})
    \end{aligned}
\end{equation}

Then the $\beta^c_{l,k}$ and $\alpha_l^c$ are obtained by,
\begin{equation}
    \begin{aligned}
        \beta^{c}_{l,k} = \frac{\text{exp}(\overline{w}^{c}_{l,k})}{\sum_{i=1}^{h} \text{exp}(\overline{w}^{c}_{l,i})}, \alpha^{c}_{l} = \frac{\text{exp}(\overline{\delta}^{c}_{l})}{\sum_{i=1}^{L} \text{exp}(\overline{\delta^{c}_{i}})}
    \end{aligned}
\end{equation}

For the sake of computational efficiency, $\alpha^c_l$, $\beta^c_{l,k}$ and $\overline{\text{C-Avg}}_c$ are calculated in advance during the data recovering for each one-class dataset $\mathcal{D}^{c}_{\text{syn}}$.

\begin{table*}[t]
    \centering
    \resizebox{\textwidth}{!}{
    \begin{tabular}{l c c c c}
    \toprule
    \multirow{2}{*}{OOD Datasets} & 
    \multicolumn{4}{c}{Softmax Score\cite{MaxConfidenceScore} / ODIN\cite{ODIN} / EBM\cite{liu2020energy} / \textbf{C2IR}(Ours)} \\
    \cline{2-5}
    &
    TNR at TPR 95\%($\uparrow$) & AUROC($\uparrow$) & Detection Acc($\uparrow$) & AUPRin($\uparrow$) \\
    \midrule
    TinyImagenet(R) & 54.79/ 82.01/ 76.17/ \textbf{88.95} & 94.92/ 95.43/ 96.37/ \textbf{98.69} & 92.46/ 88.88/ 93.22/ \textbf{96.12} & 96.62/ 94.84/ 97.52/ \textbf{98.24} \\
    TinyImagenet(C) & 63.52/ 76.79/ 81.07/ \textbf{92.37} & 95.60/ 93.68/ 96.94/ \textbf{98.45} & 92.89/ 86.59/ 93.71/ \textbf{94.98} & 97.01/ 92.61/ 97.86/ \textbf{98.03} \\
    iSUN & 55.82/ 83.43/ 76.89/ \textbf{89.19} & 94.95/ 96.08/ 96.50/ \textbf{97.87} & 92.16/ 89.65/ 93.22/ \textbf{93.24} & 96.88/ 96.11/ 97.80/ \textbf{97.94}\\
    LSUN(R) & 57.42/ 80.49/ 80.13/ \textbf{92.24} & 95.15/ 95.46/ 96.62/ \textbf{99.10} & 92.93/ 88.28/ 93.90/ \textbf{95.54} & 96.81/ 95.20/ 97.76/ \textbf{98.94} \\
    LSUN(C) & 41.95/ 74.10/ 56.55/ \textbf{86.85} & 93.52/ 89.59/ 94.84/ \textbf{98.32} & 91.98/ 84.75/ 92.20/ \textbf{93.88} & 95.78/ 85.33/ 96.50/ \textbf{98.23} \\
    SVHN & 60.33/ 56.07/ 77.59/ \textbf{94.53} & 94.26/ 87.00/ 95.58/ \textbf{98.81} & 92.93/ 80.89/ 91.61/ \textbf{94.81} & 90.55/ 71.51/ 90.39/ \textbf{96.86} \\
    CIFAR100 & 41.39/ 35.31/ 50.09/ \textbf{58.91} & 87.30/ 75.29/ 86.60/ \textbf{91.13} & 82.10/ 69.52/ 80.99/ \textbf{87.74} & 85.91/ 71.10/ 83.36/ \textbf{88.39} \\
    \hline
    Avg & 53.60/ 69.74/ 71.21/ \textbf{86.15} & 93.67/ 90.36/ 94.78/ \textbf{97.48} & 91.06/ 84.08/ 91.26/ \textbf{93.76} & 94.22/ 86.67/ 94.46/ \textbf{96.66} \\
    \bottomrule
    \end{tabular}}
    \caption{Comparison of OOD detection performance with other Post-hoc methods. We conduct our experiments on ID dataset CIFAR10 with ResNet34. For ODIN, the temperature and noise magnitude are obtained from the original training data.}
    \label{tab:OOD_main_result}
\end{table*}

\section{EXPERIMENTS}
\label{sec:experiments}
\subsection{Experimental Setup}
We use CIFAR10 as the in-distribution (ID) dataset and Resnet34 as the base model. By default, the model is first trained on natural data and fixed during test time to detect anomalous inputs from a mixture of 10000 ID images (CIFAR10 dataset) and 10000 OOD images (other natural datasets). Especially for our data-free approach, we inverted 2500 virtual samples as the labeled data at each class.


We apply some post-hoc methods as baselines: Softmax Score \cite{MaxConfidenceScore}, ODIN \cite{ODIN}, and Energy Based Method (EBM) \cite{liu2020energy}. Besides, some approaches that only depend on training data (1-D \cite{1-D}, G-ODIN \cite{baseline_G-ODIN}, Gram \cite{Gram}) and data-available reference ma-dis. \cite{ma_distance} are also considered. Following OOD literature \cite{MaxConfidenceScore, ma_distance, ODIN,liu2020energy}, we adopt the threshold-free metric \cite{AUROC} to evaluate the performance: the TNR at 95\% TPR, AUROC, Detection accuracy, and AUPRin.

\vspace{-0.8em}
\subsection{Main Results}
As shown in table \ref{tab:OOD_main_result}, we evaluate the above metrics on 7 OOD datasets. Among these baselines, Softmax Score \cite{MaxConfidenceScore} and EBM \cite{liu2020energy} (without fine-tuning) are  hyperparameter-free and the hyperparameters of ODIN \cite{ODIN} are from training data. For all OOD datasets, our C2IR outperforms other post-hoc methods without any support from natural data and, in particular, achieves the average improvement of +14.94\% TNR when 95\% CIFAR10 images are correctly classified. Moreover, our method significantly enhances the performance of the far-OOD dataset SVHN compared with near-OOD datasets (e.g., CIFAR100).

Figure \ref{fig:motivation}a and \ref{fig:motivation}b show the visualization of natural $\textit{bird}$ images in CIFAR10 and impressions from the pretrained network (ResNet34). The differences of intermediate activations means are shown in figure \ref{fig:motivation}c. As in- and out-of-distribution statistics are distinguishable at each layer, the virtual ones are closer to the natural data compared with BN running average. It demonstrates that these virtual statistics can contribute more to detecting OOD examples.
\vspace{-1.0em}

\subsection{Effect of Our OOD Detection Metric}
As shown in table \ref{tab:OOD2}, we compare C2IR with several advanced in-distribution training methods (1-D \cite{1-D}, G-ODIN \cite{baseline_G-ODIN}, GRAM \cite{Gram}) with the only access to virtual data. Besides, data-available (both ID and OOD) method Ma-dis. \cite{ma_distance} is also introduced as reference. The distribution shifts in the synthesized images cause poor performances in other baselines. Our method can utilize these virtual data to reach an ideal performance of 98.9\% average AUROC, even comparable to the data-available methods.
\begin{table}[!htbp]
    \centering
    \resizebox{8.5cm}{!}{
    \begin{tabular}{l|c|c|c|c|c}
    \toprule
    Methods & Data Access & T & L & S & Avg\\
    \midrule
    Ma dis.\cite{ma_distance} & ID \& OOD & 99.5 & 99.7 & 99.1 & 99.4  \\
    \hline
    \hline
    1-D\cite{1-D} & Virtual & 37.8 & 36.2 & 64.2 & 46.1  \\
    G-ODIN\cite{baseline_G-ODIN} & Virtual & 53.9 & 62.8 & 79.4 & 65.4  \\
    GRAM\cite{Gram} & Virtual & 84.7 & 82.5 & 94.2 & 87.1  \\
    \textbf{C2IR}(Ours) & Virtual & \textbf{98.7} & \textbf{99.1} & \textbf{98.8} & \textbf{98.9}    \\
    \bottomrule
    \end{tabular}}
    \caption{AUROC(\%) evaluation on virtual data and data-available method Ma dis. as reference. \textbf{T} denotes TinyImageNet(R), \textbf{L} denotes LSUN(R) and \textbf{S} denotes SVHN.}
    \label{tab:OOD2}
\end{table}
\vspace{-2.0em}
\subsection{Effects of MGI and Virtual Statistics}
\begin{figure}[!t]
     \centering
      \subfigure[Bird images in CIFAR10]{
      \includegraphics[width=3.6cm]{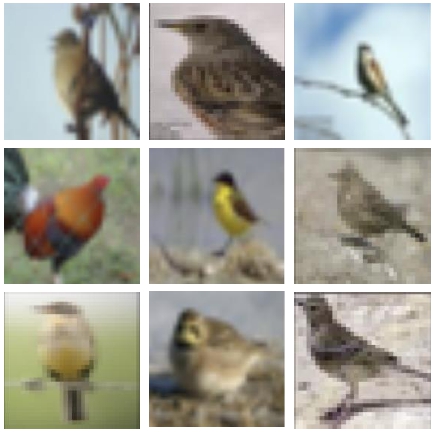}}
      \hspace{0.4em}
      \subfigure[Impressions in the network]{\includegraphics[width=3.6cm]{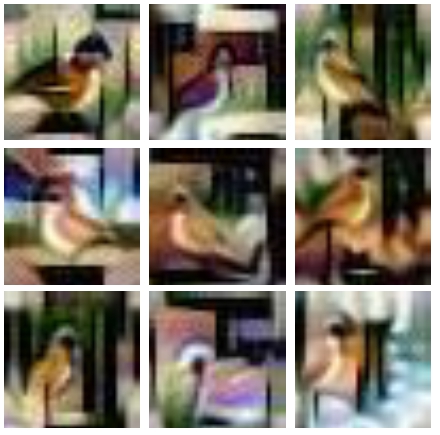}}
    \subfigure[Comparison of Act. mean]{\includegraphics[width=3.6cm]{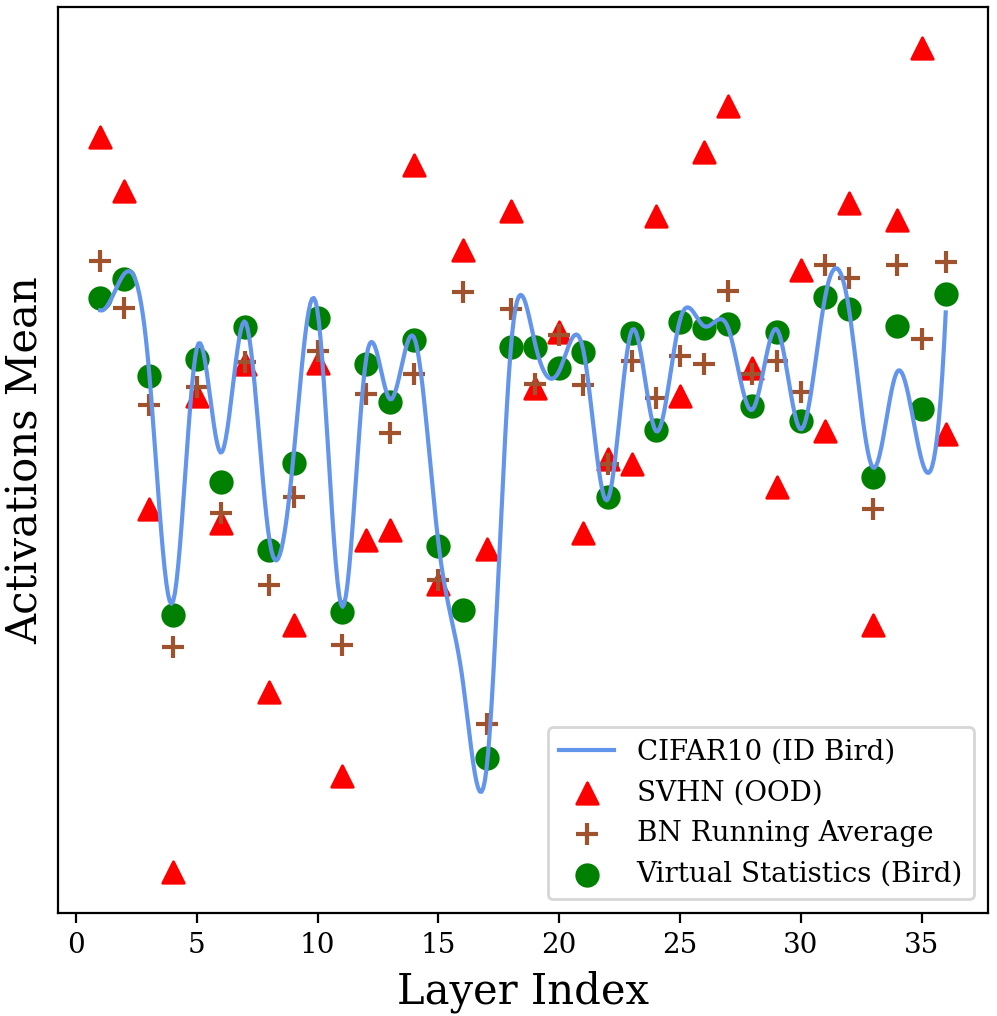}}
    \subfigure[Ablation study]{
      \includegraphics[width=3.6cm]{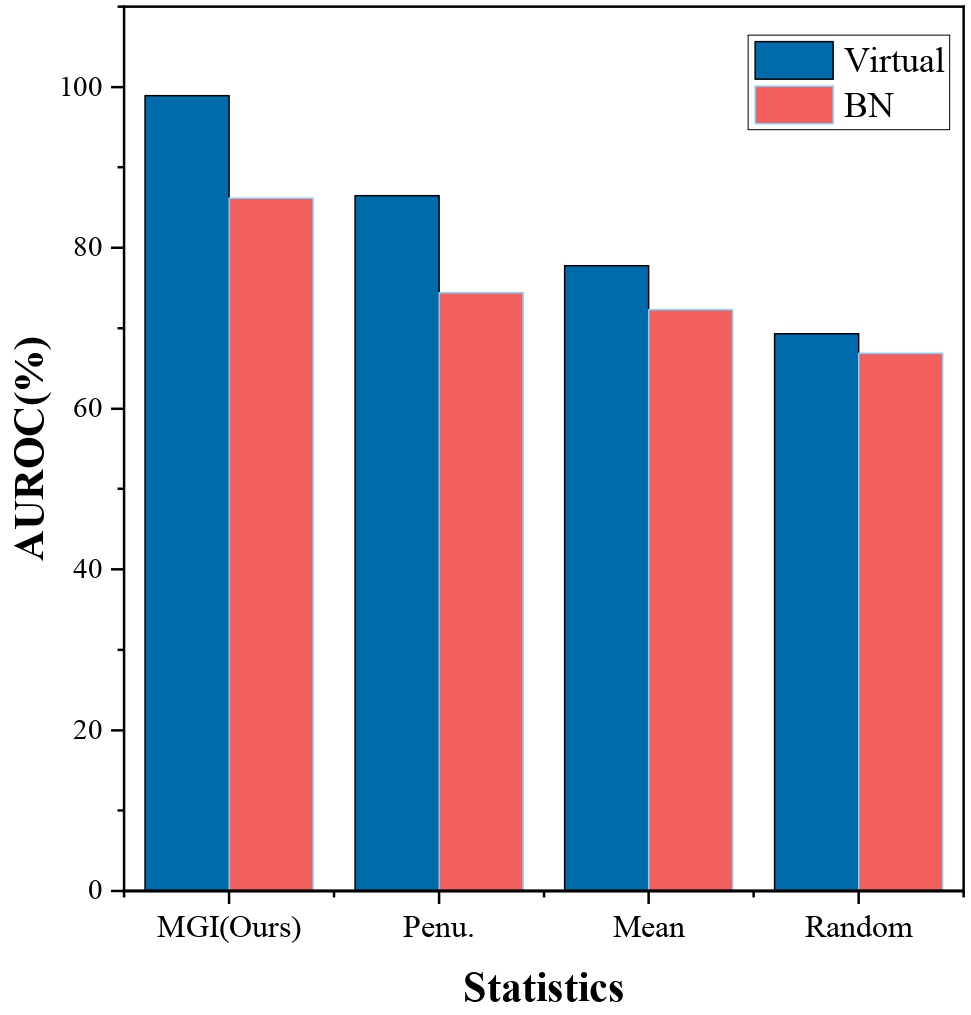}}
  \caption{\textbf{2(a)}/\textbf{2(b)}: Visualization of natural and virtual bird images. \textbf{2(c)}: We compare activations means at each layer among ID data (1000 bird images from CIFAR10), OOD data (1000 images from SVHN), virtual statistics of bird class from our method, and BN running averages. \textbf{2(d)}: Ablation study on our MGI method and virtual statistics.}
    \label{fig:motivation}
    \vspace{-1.0em}
\end{figure}

To verify the effectiveness of our MGI method and the virtual activations mean, we conduct the ablation study as figure \ref{fig:motivation}(d). MGI is compared with three baselines: 1) only use the penultimate layer (penu.) \cite{React}; 2) average all layers' statistics without weights (mean); 3) random weights. We additionally evaluate our metrics only with BN running averages for OOD detection. The results show that our virtual class-conditional activations mean outperforms BN statistics on all these strategies. Furthermore, compared with other baselines, activations mean equipped with MGI achieve the best performance.


\section{Conclusion}
\vspace{-0.5em}
We propose a novel OOD detection method without
the need for the original training dataset, which utilizes image impressions to recover class-conditional feature statistics from a fixed model. The experiments indicate that our method outperforms other post-hoc baselines with competing results. For future work, we will investigate the efficiency issues related to data preparation. Moreover, the potential of this method for continual learning is also worth exploring.


\section{ACKNOWLEDGMENT}

This work is supported by the Key Research and Development Program of Guangdong Province (grant No. 2021B0101400003) and Corresponding author is Xiaoyang Qu (quxiaoy@gmail.com).






\vfill\pagebreak

\label{sec:refs}

\bibliographystyle{IEEEbib}
\bibliography{refs}

\end{document}